\documentclass[conference]{IEEEtran}
\IEEEoverridecommandlockouts
\usepackage{cite}
\usepackage{amsmath,amssymb,amsfonts}
\usepackage{algorithmic}
\usepackage{graphicx}
\usepackage{textcomp}
\usepackage{xcolor}
\usepackage{multirow}
\usepackage{color, colortbl}
\def\BibTeX{{\rm B\kern-.05em{\sc i\kern-.025em b}\kern-.08em
    T\kern-.1667em\lower.7ex\hbox{E}\kern-.125emX}}

\definecolor{LightGray}{rgb}{0.9, 0.9, 0.9}

\begin{document}

\title{Overcoming Latency Bottlenecks in On-Device Speech Translation: A Cascaded Approach with Alignment-Based Streaming MT}

\author{\IEEEauthorblockN{Zeeshan Ahmed, Frank Seide, Niko Moritz, Ju Lin, Ruiming Xie, Simone Merello, Zhe Liu and Christian Fuegen}
\IEEEauthorblockA{\textit{Meta AI.}, USA}
}


\maketitle

\begin{abstract}

This paper tackles several challenges that arise when integrating Automatic Speech Recognition (ASR) and Machine Translation (MT) for real-time, on-device streaming speech translation. Although state-of-the-art ASR systems based on Recurrent Neural Network Transducers (RNN-T) can perform real-time transcription, achieving streaming translation in real-time remains a significant challenge. To address this issue, we propose a simultaneous translation approach that effectively balances translation quality and latency. We also investigate efficient integration of ASR and MT, leveraging linguistic cues generated by the ASR system to manage context and utilizing efficient beam-search pruning techniques such as time-out and forced finalization to maintain system's real-time factor. We apply our approach to an on-device bilingual conversational speech translation and demonstrate that our techniques outperform baselines in terms of latency and quality. Notably, our technique narrows the quality gap with non-streaming translation systems, paving the way for more accurate and efficient real-time speech translation.

\end{abstract}


\section{Introduction}

The traditional approach to speech translation involves a pipeline that combines an Automatic Speech Recognition (ASR) system with a Machine Translation (MT) system. However, current state-of-the-art cascade-based Speech Translation (ST) systems are for offline use, where the entire audio input is available in advance. In contrast, streaming setups require real-time processing of unbounded audio inputs, which poses significant challenges for cascade-based ST systems \cite{9054585, 9413492}. Conventional MT systems are the bottleneck for the real-time streaming translation set-up, as they are typically trained on sentence-aligned pairs of text. This creates a mismatch between the training data, which consists of short sentences with only a few hundred tokens, and the inference conditions (in a streaming environment), where thousands of tokens may be processed in live sessions. To address this issue, segmentation models are often used to divide the incoming text stream into manageable chunks, or segments, that can be translated independently by the MT system. However, translating the segment in isolation poses a significant risk to translation quality\cite{koehn-knowles-2017-six, cho-etal-2014-properties, pouget-abadie-etal-2014-overcoming}.

Simultaneous Machine Translation (SMT)\cite{emma2023, mma2019, milk2019, mocha2018, raffel2017online}, on the other hand, is better suited for streaming speech translation scenarios.  The SMT initiates the translation process while still receiving the input. Conventional MT relies on an encoder-decoder architecture. The decoder uses an attention module that attends to all encoder embeddings. For SMT, both the encoder and attention module must be adaptive, allowing the encoder to operate incrementally and the cross-attention in the decoder to handle variable-length encoder output sequences. In this setup, the SMT model is fed one token at a time, and at each decoding step, it decides whether it wants to wait for another token or generate a translation. The SMT systems face a trade-off between quality and latency. Prioritizing higher quality increases latency (with conventional non-streaming MT systems still representing the upper bound in quality), while reducing latency can impact translation quality. Therefore, SMT systems must carefully balance this trade-off for practical applications. Another challenge for SMT is that, despite its ability to translate input in real-time, it still requires frequent context flushing to effectively manage context and resources, particularly in on-device scenarios.

In this paper, we propose a couple of advances in cascade-based speech translation. 1) We present a novel approach to SMT called ALIgnment BAsed STReaming Machine Translation (AliBaStr-MT) which efficiently balances the quality/latency trade-off in on-device streaming speech translation. 2) We develop a robust multi-channel multi-talker ASR model that can recognize multiple languages, distinguish between different speakers and suppress side-talk/background noise. 3) Despite operating in a streaming fashion, the SMT architecture needs to keep all the previous encoder/decoder states saved in memory (kv-cached) for translation of future input. When used in an on-device setting, the memory is limited and the context needs frequent flushing. We propose incorporating segmentation cues in ASR to detect the segment boundaries and flush the states based on detected segments.


\section{Background}
\begin{figure*}[t]
    \centering
    \includegraphics[width=\linewidth, scale=0.4]{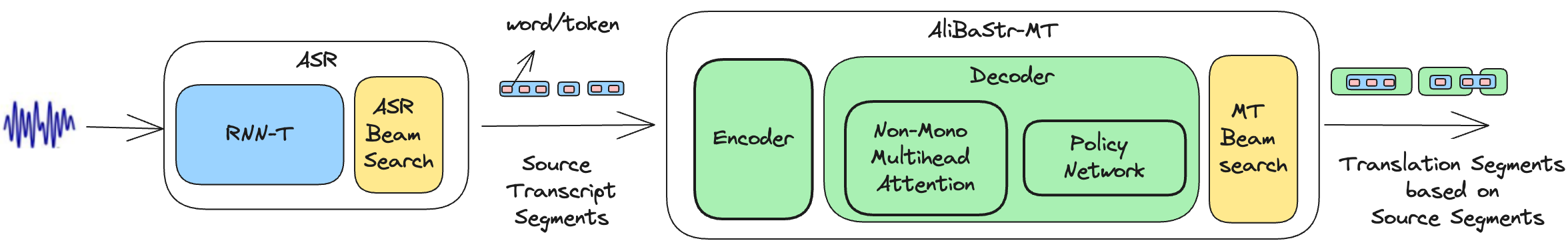}
    \caption{On-device Streaming Speech Translation Architecture using Cascade of ASR and MT. The AliBaStr-MT system processes input segments of varying sizes and dynamically decides whether to generate a translation or await additional input, guided by its read/write policy module.}
    \label{architecture}
\end{figure*}
Cascade-based and end-to-end are two major paradigms for speech translation. Cascade-based systems use separate ASR and MT models\cite{9054585, 9413492}, while end-to-end systems translate directly from source speech to target text using a single model\cite{10890381}. End-to-end systems offer advantages in terms of latency and model size, while cascade-based systems still demonstrate superior translation accuracy in many cases with well trained ASR and MT models. Furthermore, the decoupling of ASR and MT modules also allows independent training and optimization that favors low resource language pairs\cite{dabre-song-2024-nicts} for which bilingual end-to-end audio training data are scarce. Incorporating streaming capabilities into these systems is a challenging task that has received increasing attention in recent years. 

Although end-to-end models\cite{communication2023seamlessmultilingualexpressivestreaming, zhang-etal-2024-streamspeech, iranzo2021streaming, bahar2019comparative} are more favorable for streaming speech translation, cascade-based models\cite{iranzo-sanchez-etal-2024-segmentation} face considerable challenges. For instance, accurately identifying the correct segment of the ASR transcript for translation is a challenge. ASR systems often segment audio based on acoustic features such as pauses, which may not correspond to natural sentence boundaries\cite{fukuda2022speech}. This mismatch can lead to segments that are too long (spanning multiple sentences) or too short (containing only fragments), negatively affecting the accuracy of subsequent MT. To overcome this problem, researchers have explored re-segmenting ASR transcripts using punctuation restoration and language models to improve sentence boundaries and enhance MT performance\cite{fukuda2022speech, 9413432}. In this work, we incorporate the prediction of punctuation directly into the RNN-T model and employ punctuation for segment boundary detection (mostly sentence) when context flushing is needed. 


Another challenge arises when attempting to translate these segments in real-time. Conventional MT systems are designed to translate complete sentences, and are not well-suited for translating smaller segments, making them impractical for real-time streaming translation. Simultaneous MT is particularly beneficial for the real-time speech translation scenario.

SMT systems can be broadly classified into fixed policy and adaptive policy systems. In the fixed policy system, the model uses the predefined heuristics and rules to generate the translation. These heuristics remain constant throughout the translation. Fixed policy systems such as wait-$k$ or prefix-to-prefix translation systems by \cite{dalvi2018incremental, ma2019stacl} are simple and yet effective methods for streaming translation. However, the value of $k$ in wait-$k$ is not a learnable parameter of the model and is set beforehand, making it difficult to adjust based on the input and so far generated translation. The $k$ is also language dependent for different translation directions, making it difficult to build a multilingual streaming translation model with wait-$k$. \cite{guo2024glancing} on the other hand, tries to fix the quality gap in simultaneous translation models by introducing a look-ahead feature and the curriculum learning during training.

The adaptive methods, on the other hand, dynamically adjust the read/write decision based on the input and previously generated translation. The adaptive method includes the ones where non-streaming MT model is kept fixed/frozen and a heuristics or learnable module is added to make read/write decisions e.g. the confidence threshold (What-if-*) method of \cite{cho2016can} and reinforcement learning approaches by \cite{grissom2014don, gu2017learning} use a pretrained non-streaming model which is not modified during the training. \cite{press2018you} also proposed an adaptive system without attention. The model keeps the single context vector to reduce computational and memory footprint. In addition to the target tokens, the model can also generate the empty symbol to delay the translation. However, the model requires external supervision in the form of alignment between the source and target words to train the system. \cite{alinejad-etal-2021-translation} compares the full sentence translation with the partial translation and builds up a ground-truth action sequence to train the read/write module. \cite{papi-etal-2023-attention} explore the application of attention scores to create adaptive policies that guide the inference process.

Most of the recent work on adaptive policy methods is based on monotonic attention. The monotonic attention in essence learns a dynamic read/write policy which varies over the course of translation. There are various flavors available for monotonic attention, e.g. Monotonic Hard Attention \cite{raffel2017online} where the model only attends to the latest encoder state causing improvements in the latency and runtime while making a compromise on the translation quality. Monotonic Chunkwise Attention (MoChA) \cite{mocha2018} on the other hand, attends to the segment of previous encoder states to overcome the quality gap. Monotonic Infinite Lookback (MILK) \cite{milk2019} further improves quality by attending to the entire previous encoder states. These flavors can also be combined with their multi-head version called Monotonic Multi-head Attention (MMA) \cite{emma2023, mma2019}. One of the drawbacks of the MA is that the read/write policy module is tightly embedded within the decoder layers. If there are $n$ decoder layers, there will be $n$ policy modules. Whereas, only the last policy module is used for the decision.

\section{Method}

Figure \ref{architecture} depicts the overall architecture of our system. The system is composed of an streaming RNN-T based multi-channel multi-lingual ASR model with a beam-search module to generate the ASR transcripts. Translations for ASR transcripts are generated using the AliBaStr-MT system based on encoder/decoder architecture with an SMT specific beam-search component.

\subsection{Automatic Speech Recognition (ASR)}

Our ASR system is similar to \cite{lin2023directional, lin2024agadir} which is based on RNN-T. This architecture is composed of three main components: an encoder, a prediction network, and a joiner network. The primary objective of the transducer model is to generate a sequence of labels $Y = (y_1,...,y_L)$ of length $L$. These labels can represent words or word pieces, derived from an input sequence $X=(x_1,...,x_S)$ of length S, which typically consists of acoustic features such as Mel-spectral energies.

Our ASR model is designed to effectively capture and transcribe conversations in various settings. It utilizes a multi-channel approach, leveraging microphone arrays placed at different locations on the device to robustly identify the change in speaker, distinguish between conversation partners' speech and suppress side-talk. Additionally, the model is streaming multi-lingual, automatically detecting language changes within an audio stream, whether from the same or multiple speakers, and generating corresponding transcripts. This functionality is achieved through serialized output training (SOT) and a multichannel front-end, as described in \cite{lin2023directional}. During training, the RNN-T model is also trained to emit four punctuation marks (`,', `?', `!', `.') to identify segment boundaries. Finally, we employ beam search decoding on top of RNN-T to generate accurate ASR transcriptions, which are then fed into the SMT system for translation.



\subsection{Simultaneous MT}
\begin{figure}
    \centering
    \includegraphics[width=\linewidth, scale=0.4]{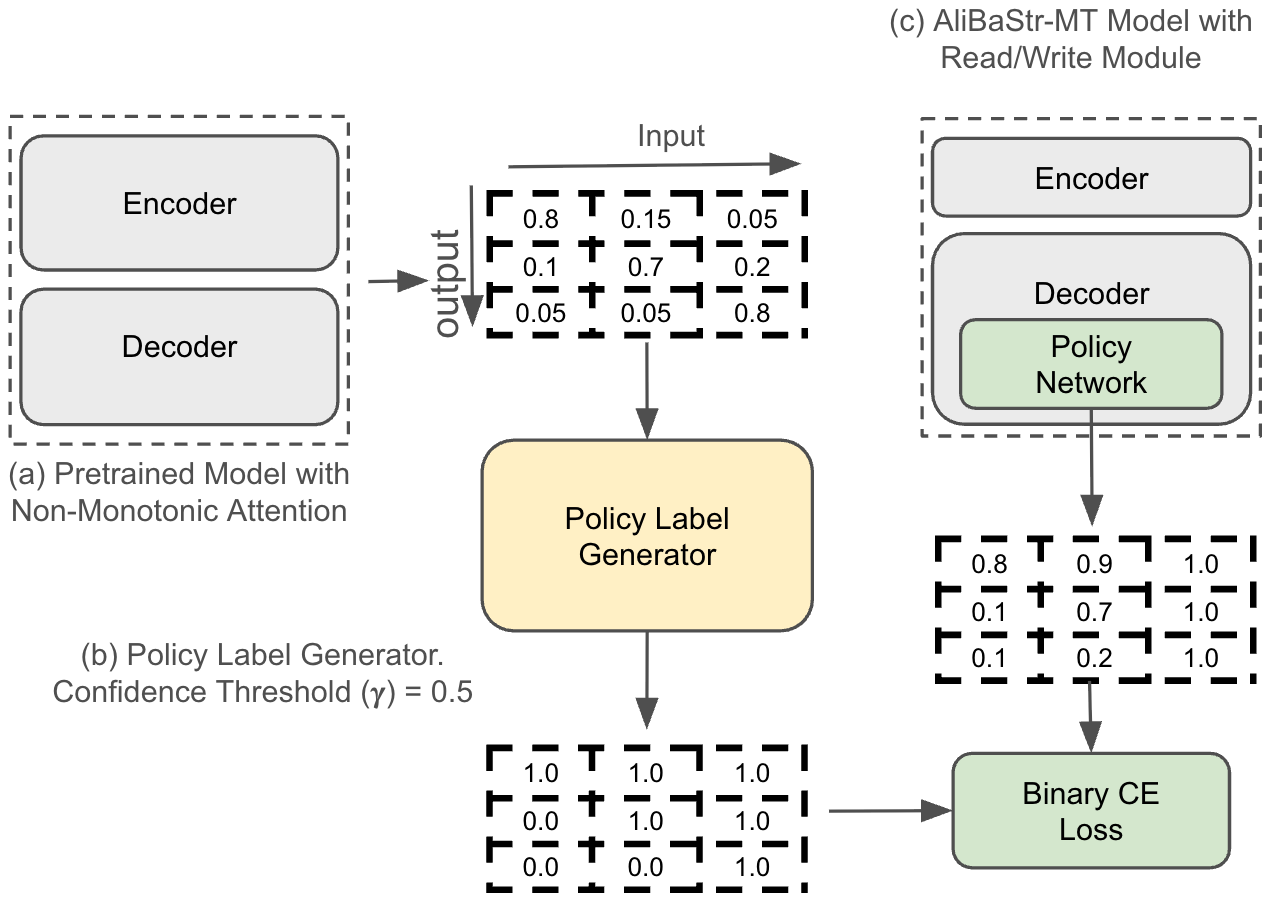}
    \caption{Read/Write Policy learning in AliBaStr-MT using a pretrained non-streaming encoder-decoder model.}
    \label{architecture_alibastr-mt}
\end{figure}

We enable streaming translation ability in a pretrained non-streaming encoder/decoder model using an additional read/write policy module. Given that the quality of the non-streaming model is currently the upper bound for the streaming model, we propose to learn the read/write policy using the alignment computed from the non-monotonic attention in a supervised manner. We get these supervised labels called pseudo-labels automatically from attention weights. The overall architecture of our ALIgnment BAsed STReaming Machine Translation (AliBaStr-MT) is shown in Figure \ref{architecture_alibastr-mt}. 


\subsubsection{Non-Streaming Full Sentence Translation Model}
The non-streaming model is trained on full sentence context using the following formulation \cite{vaswani2017attention}.  Formally, a model based on standard encoder-decoder architecture transforms an input $x = \{x_1,..,x_{|x|}\}$ into an output sequence $y=\{y_1,...,y_{|y|}\}$ using the following
\begin{align}
    h_j &= Encoder(x_i) \label{enc_eq} \\
    s_i &= Decoder(y_{i-1}, s_{i-1}, c_i) \label{dec_eq}  \\
    y_i &= Output(s_i, c_i) \label{enc_dec_out_eq}
\end{align}
The context vector $c_i$ is computed using a soft attention as follows.
\begin{align}
    e_{i,j} &= Energy(s_{i-1}, h_j) \label{energy_eq} \\
    \alpha_{i,j} &= \frac{exp(e_{i,j})}{\sum_{k=1}^T exp(e_{i,k})} \label{alpha_cxt_vec_eq}  \\
    c_i &= \sum_{j=1}^{|x|}  \alpha_{i,j} h_j \label{cxt_vec_eq}
\end{align}
The attention component of the decoder (termed as non-monotonic attention) in this architecture attends to all the encoder states $h_j : 1 \leq j \leq |x|$  for every decoding time step $ 1 \leq i \leq |y|$. The model is trained by minimizing the negative log likelihood between predicted and reference translation tokens from the training set. 

\subsubsection{Monotonic Attention and Policy Network}
\cite{emma2023, mma2019, milk2019, mocha2018, raffel2017online} proposed different types of monotonic attention. In terms of architecture, the monotonic attention used in AliBaStr-MT is similar to the non-monotonic attention except that it is equipped with a read/write module, which indicates to the model either read further input or attend to the available encoder states and generate translation.  In monotonic attention, the read/write module models the Bernoulli distribution.
\begin{align}
     e_{i,j} &= MonotonicEnergy(s_{i-1}, h_j) \label{mono_energy_eq} \\
     p_{i,j} &= Sigmoid(e_{i,j}) \label{mono_sigmoid_eq} \\
     z_{i,j} &\approx Bernoulli(p_{i,j}) 
\end{align}
If $z_{i,j} = 0$, the model reads the input and updates the encoder state, $j$ is incremented while $i$ and decoder states remain unchanged; if $z_{i,j} = 1$, the model writes the translation, updates the decoder states and increments the decoder step $i$ in addition to updating encoder states and encoder step $j$.  The training of monotonic attention involves the following
\begin{itemize}
    \item Generate the $p_{i,j}$ for all $1 \leq i \leq |y|$ and $1 \leq j \leq |x|$. 
    \item Compute the expected alignment using the $p_{i,j}$ ($\alpha_{i,j}$ and $\beta_{i,j}$ as presented in \cite{milk2019}).
    \item  Compute the context vector $c_i$.
    \item  Output translation using $c_i$.
\end{itemize}
Here, the  $p_{i,j}$ are computed first followed by the attention and output projection. In our approach presented in this paper, we follow the standard seq-to-seq training (eq. \ref{enc_eq}-\ref{cxt_vec_eq}).  Our proposed model, as shown in Figure \ref{architecture_alibastr-mt}(c), is also similar to the non-streaming model of Figure \ref{architecture_alibastr-mt}(a) except that the decoder has a read/write module making the non-monotonic attention behave as the monotonic attention. The attention module only looks at the input tokens consumed so far. It does not have access to the future input tokens (using a mask as described in section \ref{read_write_section}). The policy network is simply a binary classifier over the encoder and decoder states that predicts the probability based on the similarity between the embeddings.

\subsubsection{Read/Write Policy Learning}
\label{read_write_section}
The architecture of the AliBaStr-MT read/write module is similar to \cite{mma2019, emma2023} except that the read/write module is not part of the decoder layer. The read/write module is added on top of the decoder after all the decoder layers. We compute the current decoder state at step $i$ and compute the energy function on the current decoder state.
\begin{align}
     e^{'}_{i,j} &= MonotonicEnergy(s_{i}, h_j) \label{mono_energy_eq2} \\
     p^{'}_{i,j} &= Sigmoid(e^{'}_{i,j}) \label{mono_sigmoid_eq2} \\
     z^{'}_{i,j} &\approx Bernoulli(p^{'}_{i,j}) 
\end{align}


\begin{figure}
\centering
\includegraphics[scale=0.35]{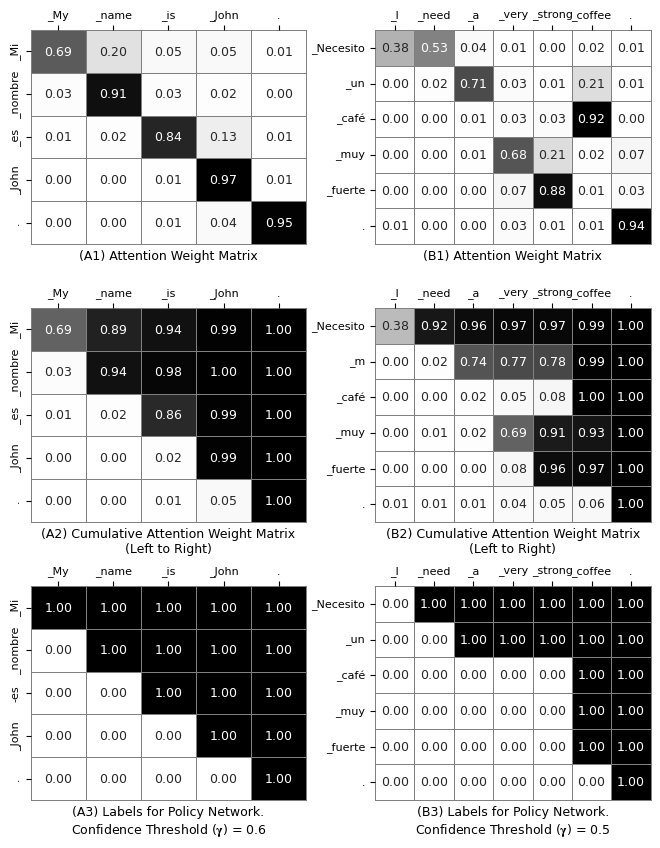} 
\caption{Conversion of Attention weight matrix to Policy label matrix for supervised training of policy network.}
\label{policy_label_matrix_ex_1}
\end{figure}

For training, the labels are generated by the Policy Label Generator (Figure \ref{architecture_alibastr-mt} (b)) which uses the attention scores from the pretrained non-streaming model. If we are able to find the point in the attention weight matrix (with some probability threshold $<$ 1) where all the input tokens have been consumed that are required to generate the $i_{th}$ output token, we can construct a label matrix for direct training of the policy network. For example, consider the input English sentence in Figure\ref{policy_label_matrix_ex_1}(A) which is translated in Spanish (using the pretrained non-streaming MT model). 
If we consider the attention weight matrix of this sentence as shown in the following Figure \ref{policy_label_matrix_ex_1}(a1), we can observe that the soft alignment points for each output token can be converted to a policy label matrix using a cumulative attention threshold (Figure \ref{policy_label_matrix_ex_1}(a2)). The policy label matrix (Figure \ref{policy_label_matrix_ex_1}(a3)) can then be used in the supervised training of the policy network. This example shows a 1:1 mapping of input tokens and output tokens. Figures \ref{policy_label_matrix_ex_1}(b) show a more complex example.

\subsubsection{Cumulative Attention Confidence Threshold ($\gamma$)}
\label{cum_attn_conf_thresh}
To convert the cumulative attention weight matrix into a discreet policy label matrix, we use an attention confidence threshold ($\gamma$). $\gamma$ ranges from 0 to 1. $\gamma$=1 means a non-streaming case where the policy label generator does not generate a "write" label until it has seen a full sentence. Any value $\gamma < 1$ enables generation of translation with fewer input tokens. The “1” in the label matrix at position (i, j) indicates a point where the model can generate the $y_{i}$ output token after reading $x_{1..j}$ input tokens.  If $\alpha_{i,j}$ represents the attention weights in the attention matrix $A$, the label matrix $L$ has the following form
\begin{align}
F &= cumsum(A, dim=1) \\
L &= genlabel(F)
\end{align}
where, 
\begin{align}
l_{i,j} &= \begin{cases}
1 & f_{i,j} \geq \gamma \\
0 & Otherwise
\end{cases} \label{eq_l_ij}
\end{align}
The policy label generator makes sure that the label matrix is monotonic in terms of input/output by enforcing the following. 
\begin{align}
l_{i,j} &= \begin{cases}
0 & l_{i-1,j} = 0 \\
l_{i, j} & Otherwise
\end{cases}
\end{align}
During training, the output $y_{i}$ is not generated with fewer input tokens than $y_{i-1}$.  For the example in Figure \ref{policy_label_matrix_ex_1}(A), $\gamma$=0.6 was used to generate the policy label matrix. However, $\gamma$=0.6 is pretty specific to this example. The example in Figure \ref{policy_label_matrix_ex_1}(B) shows $\gamma=0.5$ is a good choice.


Furthermore, there is a possibility that the highest $\alpha_{i,j}$ is not covered by the confidence
threshold ($\gamma$) given that $\gamma < 1$. The Policy Label Generator makes sure that the $write$ action is not generated until the highest $\alpha_{i,j}$ has been observed. This is achieved by adding a constraint to eq. \ref{eq_l_ij} as follows
\begin{align}
l_{i,j} &= \begin{cases}
1 & f_{i,j} \geq \gamma \:\&\: \underset{j^{'}}{\arg \max}\: (\alpha_{i,j^{'}}) \leq j\\
0 & Otherwise
\end{cases} \label{modified_eq_l_ij}
\end{align}
The read/write module operates as a binary classifier, relying on a calibration threshold ($\delta$) to convert probability scores into clear read/write decisions. Using $\delta$ instead of the training-time parameter $\gamma$, we can dynamically adjust the quality-latency trade-off during inference, making it easier to identify the optimal model configuration.

\subsection{ASR and MT Integration}
\label{str_beam_search}
To achieve low-latency and high-quality translation, it is essential to carefully integrate the ASR with AliBaStr-MT. Both ASR and MT models in our architecture are designed for real-time processing, allowing them to handle streaming data. The ASR model employs beam search to generate transcripts, which can introduce latency due to its tendency to delay output until a hypothesis is finalized. To mitigate this issue, we have implemented a design that translates both PARTIAL and FINAL hypotheses generated by the beam search.

The beam search produces the top-performing hypothesis as PARTIAL at regular intervals, while FINAL hypothesis is generated when all beams converge on a common prefix. We utilize the PARTIAL hypotheses solely for displaying translations on the screen, whereas FINAL hypotheses are used for playback during conversations.
To prevent excessive latency, we have introduced a mechanism that force-finalizes the beams if they fail to produce a FINAL hypothesis within a predetermined time frame (1.5 seconds in our experiments). 

We also employ a beam search for AliBaStr-MT. Unlike previous studies on SMT (\cite{emma2023, mma2019, milk2019, mocha2018, raffel2017online}), which relied solely on greedy decoding, our approach incorporates streaming beam search decoding. This implementation is based on \cite{rabatin2024navigating}. Similarly to the ASR, we have introduced force finalization logic into the MT beam search to reduce latency due to finalization delay. Specifically, if hypotheses are not finalized after a predetermined number of output tokens, the beams are automatically finalized.

In our streaming translation design, punctuation prediction is an integral component of the RNN-T model. The AliBaStr-MT model can process input word-by-word, generating translations based on the read/write module. In particular, it resets its internal states when it predicts the end of a sentence, denoted by the punctuation (.!?). In contrast, the non-streaming baseline model requires a complete input sentence to produce accurate translations. If only a partial sentence or short phrase is provided, the quality of the translation is compromised. When integrated with ASR, the non-streaming MT model translates the FINAL hypothesis only when it corresponds to a full sentence. However, it can also translate PARTIAL hypotheses, which may consist of shorter fragments. Since these PARTIAL translations are intended for display purposes only, some compromise in quality is acceptable for PARTIALs.

\section{Experiments}
We evaluate the performance of our speech-to-text translation approach directly on-device on English-Spanish, English-French, English-Italian directions. These language pairs offer more simultaneous translation opportunities than any other language pair.  We evaluate the models' quality/runtime and present the results for both directions for each pair, i.e. English$\rightarrow$X and X$\rightarrow$English. For evaluation, we choose internal real-life conversation (RealConv) and publicly available FLEURS \cite{conneau2023fleurs} datasets. 

\subsection{ASR Model Training}
We follow a similar approach to \cite{lin2023directional} for training our ASR system.  We use an internal dataset of de-identified video data. 
The dataset consists of 40k hours of audio. For training, we normalize the training transcripts i.e., all the transcripts are lower-cased, numbers/acronyms are verbalized and all punctuations are removed except for (`,', `?', `!', `.'). We train a bilingual ASR model for each language pair for on-device usage specifically. The ASR model is not only capable of transcribing the audio but can also label the speaker(self/other), can recognize partially overlapping speech, and perform side talk supersession.

\subsection{MT Model Training}
For training of the MT models, we use publicly available datasets, e.g. NLLB, Opensubtitle, CCMatrix, Wikipedia \cite{schwenk2019ccmatrix, fan2021beyond, lison2016opensubtitles2016, wolk2014building} etc. as well as some internal datasets. Our MT models are for on-device usage. To save memory and other resources on-device, the model for each language pair is bilingual and multitasking, serving different translation use cases. The model translates into one of two target languages; translates from ASR output where formatting is missing compared to regular text; supports code switching between languages; acts as a cross-lingual as well as same language Punc/Cap/ITN module.

\textbf{Bilingual Modeling}: To train bilingual models, a target language $identity$ token is added as a prefix to input/output to request the desired language in output. The model is then trained using a mix of English$\rightarrow$X and X$\rightarrow$English examples in the training set.

\textbf{Surface/ASR Form:} Typically, MT models are trained on written text that is properly formatted with punctuation, capitalization, and numerals, known as Surface Form. However, the output from ASR systems, referred to as ASR Form, often lacks formatting, with missing punctuation, lowercase letters, and verbalized numerals. To improve the integration of our MT model with the ASR, we apply a text normalizer to a portion of the training set, converting it into an ASR-like format on the source side. By combining this modified dataset with the original Surface Form training set, we not only improve the model's ability to translate from ASR output but also introduce cross-lingual capabilities for punctuation, capitalization, and ITN (Inverse Text Normalization).

\textbf{Copy Through:} We also augment a portion of the training set to perform an identity or a copy-through operation, i.e. English$\rightarrow$English or X$\rightarrow$X. This is to allow the model to perform code-switching when confronted with mixed language on the input. If the source side of the copy-through dataset is also text normalized, we get the same language Punc/Cap/ITN capabilities useful for the code-switching scenario.

Our primary baseline is a cascade ST system with non-streaming full sentence MT which is an Encoder/Decoder model with monotonic encoder and non-monotonic decoder having 102M parameters. However, for comparison with AliBaStr-MT, we chose a couple other state-of-the-art SMT models. \textbf{Wait-k}: Encoder/Decoder model with monotonic encoder and decoder with wait-$k$ policy\cite{ma2019stacl}. The model has around 115M parameters. \textbf{EMMA}: Efficient implementation of MMA approach \cite{emma2023, mma2019} having 124M parameters.


All of the models have the same configuration except for the architecture specific differences like presence of the tiny read/write module etc. The encoder is based on transformer architecture having 12-layers. The decoder is also transformer-based and has 5-layers. The AliBaStr-MT model has around 103M parameters. The model is similar to the non-streaming model except that it has a small read/write module. For AliBaStr-MT, we set $\gamma=0.5$ for policy label generation and adjust $\delta$ during inference to control quality-latency trade-off. For evaluation, we use our streaming beam-search implementation as introduced in section \ref{str_beam_search}. The non-streaming models use the standard beam search. For real-time streaming translation, the ASR output (FINAL/PARTIAL) is immediately sent to the SMT model for translation except for the last token which is held for the next call. This is done due to the difference in the ASR and MT sentencepiece vocabulary. Finally, we measure the quality of the model using BLEU, and the latency of the streaming models using Average Lag (AL)\cite{ma2019stacl}. We also introduce User Perceived Latency (UPL), which measures the time it takes for the system to produce the first or last output token after receiving the initial or final input audio frame. As its name implies, UPL is directly observable by users, who can easily notice delays at the beginning and end of the sentence. However, detecting latency during the intermediate stages of translation is more challenging. AL presented in the results section is non-computational-aware while UPL compensates for the computational-aware AL.

\subsection{Results}
Table \ref{asr_results_table} provides an overview of our ASR system's performance, highlighting the Word Error Rate (WER) for each language pair (English-X). The results are presented for two distinct datasets: RealConv and FLEURS. Notably, the table also includes Speaker Attribution (SA) errors for the RealConv dataset, which measures the accuracy of predicting speaker tags in multi-party speech recognition scenarios.
\begin{table}
\centering
\caption{Word Error Rate (WER) and Speaker Attribution Error (SA) of the ASR System on English$-$X pairs.}
\setlength{\tabcolsep}{3pt}
\begin{tabular}{|lccc|cc|}
\hline
& \multicolumn{3}{c|}{\textbf{Real-life Conversation}} & \multicolumn{2}{c|}{\textbf{FLEURS}}\\
\textbf{Language} & $\downarrow$WER (X) & $\downarrow$WER (EN) & $\downarrow$SA & $\downarrow$WER (X) & $\downarrow$WER (EN)\\
\hline
\hline
EN-ES & 11.99 & 7.55 & 0.01 & 10.44 & 4.95\\
EN-FR & 16.61 & 7.73 & 0.07 & 13.90	& 10.21\\
EN-IT & 15.99 & 8.08 & 0.05 & 10.56 & 5.89\\
\hline
\end{tabular}
\label{asr_results_table}
\end{table}

\begin{table}
\centering
\caption{BLEU and Latency (P50) metrics of ST models for  English$\leftrightarrow$X directions on Real-life Conversation dataset.}
\begin{tabular}{|llcc|c|cc|}
\hline
& & \multicolumn{2}{c|}{} & \multicolumn{3}{c|}{\textbf{$\downarrow$Latency}}\\
\cline{5-7} 
& & \multicolumn{2}{c|}{\textbf{$\uparrow$BLEU}} & \multirow{2}{*}{\textbf{AL}} & \multicolumn{2}{c|}{\textbf{UPL}}\\
\cline{6-7} 
\textbf{X} & \textbf{Model} & X$\rightarrow$EN & EN$\rightarrow$X & & First & Last\\
\hline
\hline
\rowcolor{LightGray}
\multirow{4}{*}{ES} & Non-Streaming & 45.56 & 46.60 & 4.39 & 4.81 & 1.99 \\
 & Wait-$4$ & 42.37 & 40.77 & 3.34  & 3.46 & 2.18 \\
 & EMMA & 40.84 & 41.71 & 3.20 & 3.19 & 2.32\\
 & AliBaStr-MT & \textbf{43.89} & \textbf{45.30} & \textbf{2.74} & \textbf{3.16}	& \textbf{1.36} \\
\hline
\rowcolor{LightGray}
\multirow{4}{*}{FR} & Non-Streaming & 43.02 & 46.32 & 4.15 & 4.32 & 2.07 \\
 & Wait-$4$ & 40.82 & 42.46 & 3.28 & 3.40 & 2.46\\
 & EMMA & 40.13 & 43.06 & 3.34 & \textbf{3.03} & 2.67 \\
 & AliBaStr-MT & \textbf{42.13} & \textbf{45.06} & \textbf{2.83} & 3.18 & \textbf{1.67} \\
\hline
\rowcolor{LightGray}
\multirow{4}{*}{IT} & Non-Streaming & 41.70	& 32.66 & 4.49 & 4.86 & 1.86 \\
 & Wait-$4$ & 41.18 & 31.15 & 3.36 & 3.58 & 2.06\\
 & EMMA & 38.56 & 30.85 & 3.60 & \textbf{3.40} & 2.44 \\
 & AliBaStr-MT & \textbf{40.99} & \textbf{32.41} & 2.89 & 3.43 & \textbf{1.47}\\
\hline
\end{tabular}
\label{mt_results_table}
\end{table}

Table \ref{mt_results_table} presents the end-to-end speech-to-text translation results using a beam size of 1. For each pair of languages, the table reports BLEU for both directions together with AL and UPL in seconds. The scores presented for AL and UPL are the average for both directions. AL differs from UPL in the sense that AL accounts for the amount of speech (in seconds) taken to generate the output token, while UPL measures the wall-clock time taken to predict the first (last) output token since the system consumed the first (last) input audio frame. 

Table \ref{mt_results_table} compares the performance of four different models. Our target baseline is, however, the non-streaming model which is the best in quality and has the BLEU that is an upper bound for all the streaming models. In contrast, the latency of the non-streaming model is the worst. On the other hand, the latency of AliBaStr-MT is best across all models. Both non-computational-aware AL and UPL show significant gains compared to the non-streaming model. AliBaStr-MT also has better translation quality (measured in terms of BLEU) than the Wait-$k$ and EMMA models and approaches the quality of the non-streaming model.


\begin{figure}
    \centering
    \includegraphics[width=\linewidth]{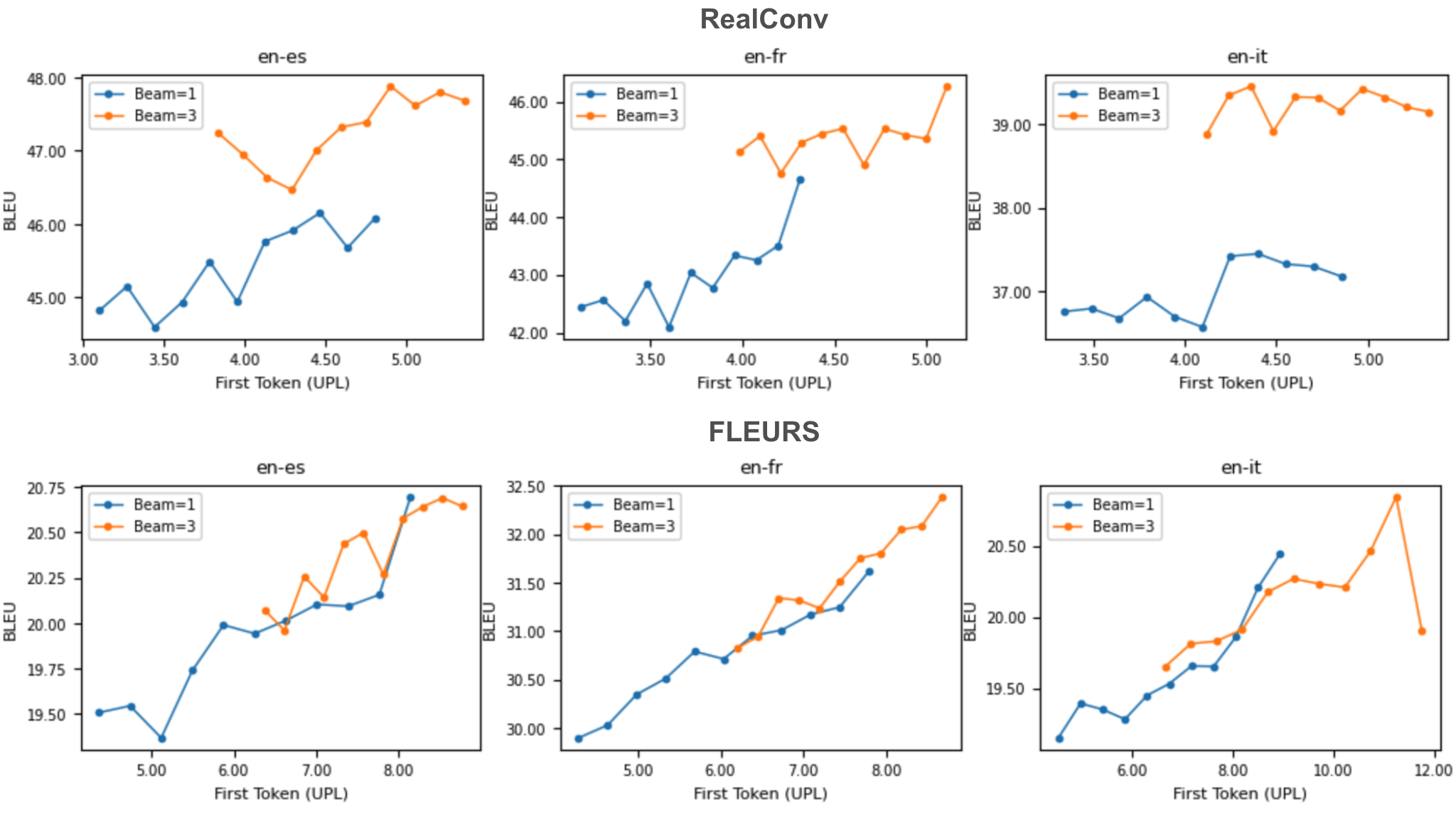}
    \caption{First Token (UPL) vs. BLEU on RealConv and FLEURS datasets.}
    \label{fig:bleu_upl_alibastr}
\end{figure}

Furthermore, Figure \ref{fig:bleu_upl_alibastr} presents the latency-quality trade-off curve using the AliBaStr-MT model. These trend curves are generated using $\delta=(0.50, 0.55, 0.60, ..., 1.0)$. For each pair of languages, the graph presents the trend at beam sizes of 1 and 3. For beam size $>1$, we kept the force finalization token limit to 5 i.e., whenever a beam has 5 or more output tokens, the beams are finalized and only the highest scoring beam that has the maximum number of output tokens is retained. As expected, there is a slight difference in BLEU at beam sizes of 1 and 3. However, with increasing beam size, the model latency also increases (note that this difference is not captured by the non computation-aware AL which only measures the amount of speech taken to generate output token). In addition to the RealConv dataset, Figure \ref{fig:bleu_upl_alibastr} also presents the trend for the FLEURS dataset. For FLEURS dataset, the trend is similar to the RealConv dataset, as UPL increases, the BLEU increases. The highest UPL and BLEU is at $\delta=1.0$ where the AliBaStr-MT model performs similarly to the non-streaming model.




\bibliographystyle{./IEEEtran}
\bibliography{./IEEEabrv,./streaming_st}

\end{document}